\documentclass{article}

\usepackage{rl-notation}
\usepackage{multirow}
\usepackage{pdfpages}

\usepackage{microtype}
\usepackage{graphicx}

\usepackage{subcaption}
\usepackage{booktabs} 

\usepackage{hyperref}

\usepackage[accepted]{icml2023}

\usepackage{amsmath}
\usepackage{amssymb}
\usepackage{mathtools}
\usepackage{amsthm}

\usepackage[capitalize,noabbrev]{cleveref}

\theoremstyle{plain}

\theoremstyle{definition}

\theoremstyle{remark}

\def\C{{\mathrm C}}
\usepackage[textsize=tiny]{todonotes}

\icmltitlerunning{Explaining Reinforcement Learning with Shapley Values}

\begin{document}
\twocolumn[
\icmltitle{Explaining Reinforcement Learning with Shapley Values}



\icmlsetsymbol{equal}{*}

\author{Daniel Beechey\thanks{Department of Computer Science, University of Bath, UK. Correspondence to: Daniel Beechey $<$djeb20@bath.ac.uk$>$} \And Thomas M. S. Smith\samethanks \And Özgür Şimşek\samethanks}

\begin{icmlauthorlist}
\icmlauthor{Daniel Beechey}{bath}
\icmlauthor{Thomas M. S. Smith}{bath}
\icmlauthor{\"{O}zg\"{u}r \c{S}im\c{s}ek}{bath}
\end{icmlauthorlist}

\icmlaffiliation{bath}{Department of Computer Science, University of Bath, UK}

\icmlcorrespondingauthor{Daniel Beechey}{djeb20@bath.ac.uk}

\icmlkeywords{ICML, machine learning, reinforcement learning, explainable AI, Shapley values, SVERL}

\vskip 0.3in
]


\printAffiliationsAndNotice{}  

\begin{abstract}
For reinforcement learning systems to be widely adopted, their users must understand and trust them. We present a theoretical analysis of explaining reinforcement learning using Shapley values, following a principled approach from game theory for identifying the contribution of individual players to the outcome of a cooperative game. We call this general framework Shapley Values for Explaining Reinforcement Learning (SVERL). Our analysis exposes the limitations of earlier uses of Shapley values in reinforcement learning. We then develop an approach that uses Shapley values to explain agent performance. In a variety of domains, SVERL produces meaningful explanations that match and supplement human intuition.
\end{abstract}

\section{Introduction}
\label{sec:intro}
Reinforcement learning systems have potential for significant impact in real-world applications. To be widely adopted, it is useful for these systems to not only perform well but also be explainable.

Methods for explaining reinforcement learning can be categorised as \emph{intrinsically interpretable} or \emph{post-hoc}. Intrinsically interpretable approaches improve the transparency of models by substituting an opaque model with a more understandable one, such as a decision tree. This approach often leads to a reduction in representational power. In contrast, post-hoc methods hold no constraints on the complexity of the model, treating it as a black box. Reinforcement learning systems with the largest potential to positively benefit society depend on function approximators with large representational power, such as deep neural networks. We therefore focus on post-hoc explanation methods. 

An established, post-hoc explanation method for supervised learning uses Shapley values \citep{Shapley1953}, a principled approach from game theory for identifying the contribution of individual players to the outcome of a cooperative game. Shapley values are the result of a rigorous mathematical formulation that satisfies four axioms of fairness. In supervised learning, Shapley values explain a model by expressing the contribution of individual features to the predictions of the model. 

We analyse, from first principles, how Shapley values can be used to explain reinforcement learning. We make three main contributions. First, we develop a theoretical framework for using Shapley values in the context of reinforcement learning, showing that earlier uses of Shapley values in reinforcement learning are incorrect or incomplete. Secondly, we consider which aspects of reinforcement learning are important to explain, arguing that explaining agent performance is an important and overlooked element. Thirdly, we develop a principled approach that identifies the contributions of state features to the performance of an agent. 

We call this general framework Shapley Values for Explaining Reinforcement Learning (SVERL). In a variety of domains, SVERL produces meaningful explanations that match and supplement human intuition.  

\section{Background}
\label{sec:prem}
We model the interaction of an agent with its environment as a Markov Decision Process (MDP), defined by the tuple $(\S, \A, p, r, \gamma, p_0)$, where $\S$ denotes the set of states, $\A$ the set of actions, $p : \S \cross \A \cross \S \rightarrow [0, 1]$ the transition dynamics, $r : \S \cross \A \rightarrow \mathbb{R}$ the reward function, $\gamma \in [0, 1]$ the discount factor, and $p_0 : \S \rightarrow [0, 1]$ the initial state distribution. At decision stage $t$, $t \ge 0$, the agent observes the current state of the environment, $s_t \in \S$, and executes action $a_t \in \A(s_t)$. Consequently, the environment transitions to a new state, $s_{t+1} \sim p(\cdot | s_t, a_t)$, and returns reward $r_{t+1}$ whose expected value is $r(s_t, a_t)$. The objective is to learn a policy $\pi$ that maximises the expected return $\mathbb{E}_\pi[G_t]$, where $G_t = \sum_{k=t}^\infty\gamma^{k} r_{k+1}$. The policy can be stochastic, $\pi : \S \cross \A \rightarrow [0, 1]$, or deterministic, $\pi : \S \rightarrow \A$. A state-value function, $V^{\pi}{\left(s\right)}$, gives the expected return from state $s$ when following policy $\pi$, $V^{\pi}{\left(s\right)} = \mathbb{E}_\pi{\left[G_t|s_t = s\right]}$. A state-action value function, $Q^{\pi}{\left(s, a\right)}$, gives the expected return from state $s$ if the agent executes action $a$ and follows policy $\pi$ thereafter, $Q^{\pi}{\left(s, a\right)} = \mathbb{E}_\pi{\left[G_t|s_t=s, a_t=a\right]}$. The optimal state value function is denoted by $V^*$ and the optimal state-action value function by $Q^*$.

We assume that an environment has a set of $n$ state features, $\F = \{0, \dots, n-1\}$, where we can decompose the state space according to the state features, $\S = \S_0 \cross \dots \cross \S_{n-1}$, and each state can be represented as an ordered set: $s=\{s_i | s_i \in \S_i\}_{i=0}^{n-1}$.  For example, in a classic gridworld domain, a state could be the agent's location, with $x$ and $y$ coordinates as state features. Let $\C \subset \F$ be a set of observable state features. Then a partial observation of a state is the ordered set $s_\C = \{s_i | i \in \C\}$.

Shapley values assign the contributions of individual players to the outcome of a cooperative game~\citep{Shapley1953}. They are the unique solution to a set of mathematical axioms that specify fair distribution of credit across players. A cooperative game is defined by a set $\F$ of players and a characteristic value function $v : 2^{|\F|} \rightarrow \mathbb{R}$, where $v\left(\C\right)$ returns the outcome of the game when played by some coalition of players $\C \subseteq \F$, with $v(\emptyset) = 0$. The Shapley value of player $i$ in the game $(\F, v)$ is:
\begin{equation}\label{eq:sv}
	\phi_i\left(v\right) = \sum_{\C\subseteq \F\setminus \left\{i\right\}}\frac{\left|\C\right|!\left(\left|\F\right|-\left|\C\right|-1\right)!}{\left|\F\right|!}\cdot \delta\left(i,\C\right),
 \end{equation}
where $\delta\left(i,\C\right) = v\left(\C\cup\left\{i\right\}\right)-v\left(\C\right)$ is the marginal gain in characteristic value when player $i$ joins coalition $\C$. As an example, the employees of a company can be modelled as players in a game where profit is the characteristic value function. 

Shapley values have been adopted in machine learning to determine the contribution of features to the predictions of supervised learning models~\citep{Lipovetsky2001}. Let $f_\F : \X \rightarrow \Y$ be a supervised learning model defined over a set of $n$ features, $\F = \{0, \dots, n-1\}$, such that $\X = \X_0 \cross \dots \cross \X_{n - 1}$ and each $\x \in \X$ can be represented as an ordered set, $\x = \{x_i | x_i \in \X_i\}_{i=0}^{n-1}$. Then Shapley values show the contribution of feature $x_i \in \x$ to the target $y = f_\F(\x)$ for the single point $\x$. As an example, when predicting the quality of wine using features such as acidity, pH, and alcohol \citep{Cortez2009}, the Shapley values show how much each feature contributes to the predicted quality of a specific wine. This is done by modelling the prediction at $\x$ as a game, where the features $\{x_0, \dots, x_{n-1}\}$ are the players and the target prediction $y = f_\F(\x)$ is the outcome of the game. Then the Shapley values $\phi_i(f, \x)$, specifying the contribution of feature $x_i$ to the prediction $y = f(\x)$, are computed using the characteristic value function:

\begin{equation} \label{eq:Shapley valuesl}
    v^f{\left(\C\right)} \defeq f_\C(\x), \nonumber
\end{equation}

where $\C \subseteq \F$ and $f_\C(\x)$ is the model's prediction for the ordered set $\x_\C = \{x_i | i \in \C\}$. The resulting Shapley values satisfy $f_\F(\x) = v^f{\left(\emptyset\right)} + \sum_{i \in \F}\phi_i(f, \x)$.

Shapley values show each feature's contribution to the change in prediction when all features are known, $f_\F(\x)$, compared to when no features are known, $f_\emptyset(\x) = v^f{(\emptyset)}$. In game theory, the value of a game with no players is zero. Hence $v(\emptyset) = 0$. In supervised learning, the prediction when no features are known is the expected model prediction over the data distribution. Hence $v^f(\emptyset) = \mathbb{E}_{p(\x)}{\left[f(\x)\right]}$, where $p(\x)$ is the data distribution, the probability that a randomly sampled point from $\X$ equals $\x$.

Computing Shapley values requires predictions, $f_\C(\x)$, to be made for all subsets of features, $\C\subseteq \F$. The original approach to approximating such predictions was to retrain the model for all $\C \subseteq \F$ \citep{Strumbelj2009}. With a large number of features, this is infeasible. An alternative method defines the prediction at $\x$ with subset of features $\C$ as:
\begin{equation}\label{eq:offmanifoldexpectation}
    f_\C(\x) = \mathbb{E}_{p(\x')}\left[f_\F(\x_\C\cup \x'_{\bar{\C}})\right],
\end{equation}
where $p(\x')$ is the data distribution \citep{Strumbelj2010,Strumbelj2014}. \cref{eq:offmanifoldexpectation} can be approximated by marginalising over possible values for the unobserved features $\bar{\C} = \F \setminus \C$. Assuming independent features and sampling $n$ data points,
\begin{equation}\label{eq:offmanifoldsampling}
	f_\C(\x) = \lim_{n\rightarrow \infty}{\frac{1}{n}\sum_{\x' \sim p(\x')}{f_\F(\x_\C\cup\x'_{\bar{\C}})}}.
\end{equation}
Using \cref{eq:offmanifoldsampling}, an unbiased approximation algorithm for calculating Shapley values samples a marginal gain:
\begin{equation}
    \label{eq:offmanifoldapprox}
    \hat{\delta}(i, \C) = f_\F(\x_{\C\cup\{i\}}\cup\x'_{\overline{\C\cup\{i\}}})-f_\F(\x_\C\cup\x'_{\bar{\C}}),
\end{equation}
where the coalition $\C\subseteq \F\setminus \left\{i\right\}$ is sampled proportional to the multinomial term in \cref{eq:sv} and $\x' \sim p(\x')$. The mean of these samples is the Shapley value in the limit \citep{Strumbelj2010}. This algorithm does not require retraining the models. It is one of the approximations used in the popular python package SHAP \citep{Lundberg2017}, which calculates Shapley values for an arbitrary machine learning model. There are other approximations included in SHAP; they all approximate \cref{eq:offmanifoldexpectation} in some way.

\cref{eq:offmanifoldexpectation} is referred to as \emph{off-manifold}. It makes the simplifying assumption that the features are independent. When features are correlated, this assumption samples points $\x_\C\cup\x'_{\bar{\C}}$ that may not lie on the data manifold. Without this simplifying assumption, the prediction at $\x$ with subset of features $\C$ becomes:
\begin{equation}\label{eq:onmanifoldexpectation}
	f_\C(\x) = \mathbb{E}_{p(\x'|\x_\C)}\left[f_\F(\x')\right],
\end{equation}
where the conditional data distribution $p(\x'|\x_\C)$ takes into account the feature correlations~\citep{Frye2020a}. An \emph{on-manifold} sampling method that uses \cref{eq:offmanifoldapprox} but now samples $x' \sim p(\x'|\x_\C)$ can then be used to approximate Shapley values for models with correlated features.

Shapley values $\phi_i(f, \x)$ provide the local contribution of features to a prediction. The local contributions can be combined to identify the global Shapley value for a feature, producing the mean contribution of a feature to a model's predictions: $\Phi_i(f) = \mathbb{E}_{p(\x)}\left[\phi_i(f, \x)\right]$. If we consider a new characteristic value function, defined using a model's loss $\ell$, $v^{\ell}(\C) \coloneqq \ell\left(f_\emptyset(\x), y\right) - \ell\left(f_\C(\x), y\right)$, then global Shapley values can be interpreted as the contribution of feature $i$ to the model's prediction accuracy \citep{Covert2020}:
\begin{equation}
	\Phi_i(f) = \mathbb{E}_{p(\x)}\left[\phi_i(v^{\ell}, \x)\right]. \nonumber
\end{equation}

In reinforcement learning, earlier work has directly applied the SHAP package to an agent's policy \citep{Rizzo2019,Wang2020,He2020,Remman2021,Lover2021,Liessner2021} or state-value function~\citep{Zhang2020,Zhang2022} in an effort to explain reinforcement learning in specific applications. This earlier work implicitly assumes that the state features are independent because SHAP implements only off-manifold approximations. More importantly, this earlier work has not explored the theoretical basis for what the resulting Shapley values mean in the context of reinforcement learning.

In the following sections, we present a theoretical and empirical analysis of how Shapley values can be used to explain reinforcement learning, starting from first principles. We refer to this general framework as Shapley Values for Explaining Reinforcement Learning (SVERL). 

\section{Using Shapley Values to Explain Reinforcement Learning}
\label{sec:svinrl}

We start by exploring the use of Shapley values to explain the value function and the policy of an agent. Our analysis shows that (1) applying Shapley values to a value function produces explanations that have no relation to the performance or behaviour of an agent, and (2) applying Shapley values to policies explains the contribution of state features to an agent's decisions but not to its performance.

\textbf{Shapley values applied to value functions.} In order to use Shapley values to explain an agent's value function, we follow the theory of on-manifold Shapley values in supervised learning to propose the following characteristic value functions for $V$ and $Q$:
\begin{align}
	v^{\hat{V}}\left(\C\right) &\defeq \hat{V}^\pi_\C(s) = \sum_{s' \in \S}{p^{\pi}(s'|s_\C)\hat{V}^\pi(s')} \label{eq:valueshap} \\
	v^{\hat{Q}}\left(\C\right) &\defeq \hat{Q}^\pi_\C(s, a) = \sum_{s' \in \S}{p^{\pi}(s'|s_\C)\hat{Q}^\pi(s', a)} \label{eq:qvalueshap}
\end{align}
\cref{eq:valueshap,eq:qvalueshap} account for feature correlations by using the conditional limiting state occupancy distribution $p^{\pi}(s'|s_\C)$, the probability of being in state $s'$ given that $s_\C$ is observed and the agent is following policy $\pi$.

Shapley values resulting from \cref{eq:valueshap} satisfy $v^{\hat{V}}\left(\F\right) = v^{\hat{V}}\left(\emptyset\right) + \sum_{i \in \F}\phi_i(v^{\hat{V}}, s)$. They show each feature's contribution to the change in characteristic value when all state features are observed, $\hat{V}^\pi(s)$, compared to when no state features are observed, $\hat{V}^\pi_{\emptyset}(s)$. This observation also holds for \cref{eq:qvalueshap} and all other characteristic value functions for reinforcement learning presented in this paper.

One might expect the Shapley values resulting from \cref{eq:valueshap,eq:qvalueshap} to relate to performance in some way, given that a value function represents an agent's prediction of how well its policy performs. However, these characteristic value functions refer to the expected return of the agent's \emph{original} policy $\pi$. Not observing a state feature is likely to change an agent's policy, which in turn changes the expected return. By never evaluating any change in policy, the full consequences of removing state features are not being considered. Consequently, the resulting explanations do not meaningfully relate to performance or behaviour.

Instead, Shapley values applied to the value function explain the contribution of each feature to the \emph{predictions} of the value function\textemdash but only under the assumption that all features will be observed by the agent when acting in the environment. This is a subtle but important point. Shapley values applied to the value function do not explain the agent's performance; they explain the value function as a predictor---but without considering the impact of features on behaviour.

\begin{figure}[!tb]
    \centering
    \graphicspath{{./plots/}}
    \begin{subfigure}[b]{0.32\columnwidth}
        \centering
        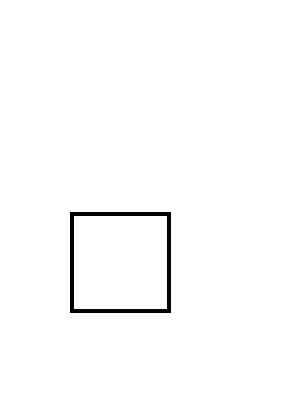    
        \caption{Gridworld-A}
        \label{fig:gwa}
    \end{subfigure}
    \begin{subfigure}[b]{0.32\columnwidth}
        \centering
        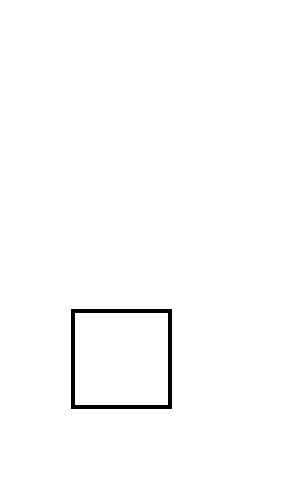    
        \caption{Gridworld-B}
        \label{fig:gwb}
    \end{subfigure}
    \begin{subfigure}[b]{0.32\columnwidth}
        \centering
        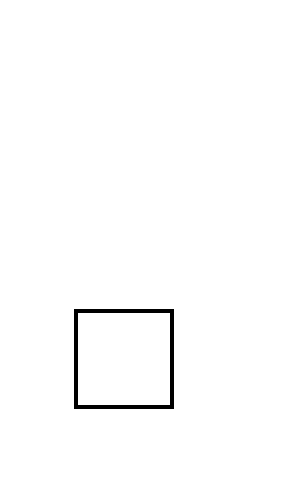    
        \caption{Gridworld-C}
        \label{fig:gwc}
    \end{subfigure}
    \caption{Deterministic gridworlds, with actions North, South, East, and West. The numbers in each grid square show the state identifier. The initial state is either state $1$ or state $2$, with equal probability. The reward is $-1$ for each action and an additional $+10$ for transitions into a terminal state (G). The discount factor $\gamma$ is $1$. State features are the $x$ and $y$ coordinates. The red arrows show the optimal action in each state.}
    \label{fig:gws}
\end{figure}

We use two examples to illustrate the difference between explaining the value function as a predictor and explaining agent performance. We use \cref{eq:qvalueshap} to apply Shapley values to $Q^*$ in Gridworld-A, shown in \cref{fig:gwa}, and \cref{eq:valueshap} to apply Shapley values to $V^*$ in Tic-Tac-Toe.

\begin{figure}[t]
    \centering
    \includegraphics[width=3.2in]{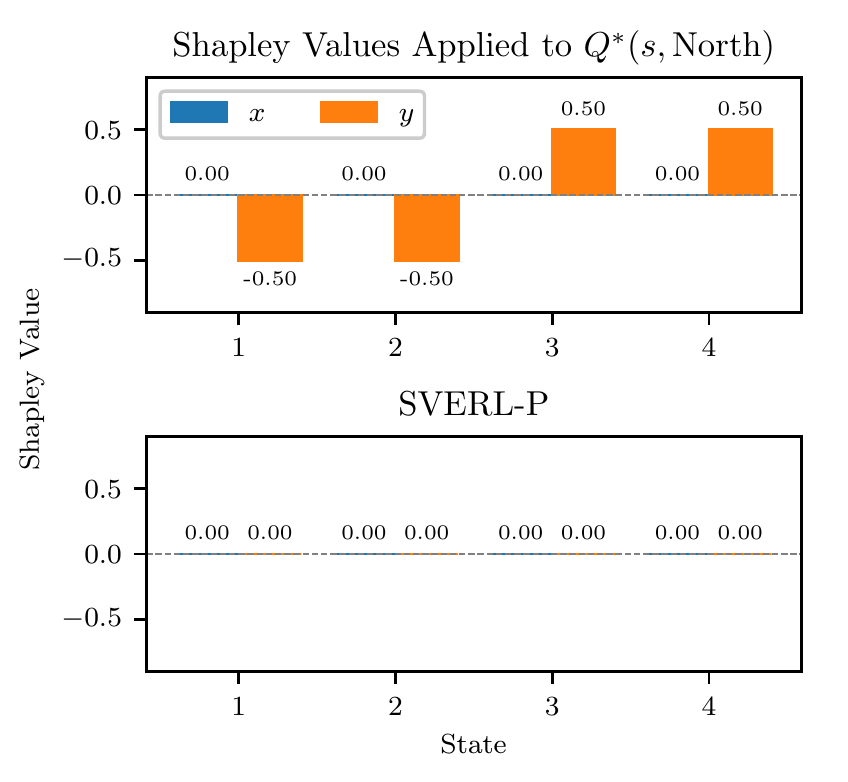}
    \caption{Top panel: Shapley values applied to a state-action value function in Gridworld-A (\cref{fig:gwa}). Bottom panel: SVERL-P, presented in \cref{sec:SVERL-P}, for the same domain. The optimal action is always North so we intuitively expect contributions to performance to be zero for all state features in all states. This is accurately captured by SVERL-P but not by Shapley values applied to the value function. 
    }
    \label{fig:value_comparison}
\end{figure}

In Gridworld-A, the optimal action is North (N) in each state. Intuitively, if the optimal action is the same in all states, then the contribution of each state feature to performance should be zero. However, Shapley values applied to $Q^*(s, N)$, shown in \cref{fig:value_comparison} (top panel), produce non-zero contributions for the $y$ state feature.

To explore why, consider the contribution of $y$ in state 1. If neither $x$ nor $y$ is known, the agent is equally likely to be in states 1, 2, 3, or 4 (we are ignoring terminal states), with $Q^*(s, N)$ values of 8, 8, 9, and 9, respectively. Consequently, the predicted return is 8.5. Now consider the marginal gain from observing $y$. If $y$ is known to be 1 and $x$ remains unknown, the agent is equally likely to be in states 1 and 2, with $Q^*(s, N) = 8$ for both states, yielding a predicted return of 8. Hence, the marginal gain in prediction from observing $y$ is $8 - 8.5 = -0.5$. Similarly, if $x$ is known to be 1 and $y$ is unknown, the agent is equally likely to be in states 1 and 3, with $Q^*(s, N)$ values of 8 and 9, respectively, yielding a predicted return of 8.5. If $y$ is also known, then predicted return is $Q^*(1, N) = 8$. Hence, the marginal gain in prediction when observing $y$ is again $8 - 8.5 = -0.5$. Both marginal gains are $-0.5$, resulting in a Shapley value of $-0.5$ for $y$ in state 1.

In contrast, the \emph{actual} return from state 1 is 8, whatever combination of features is observed, because the optimal policy selects North in every state. Human intuition therefore assigns a contribution of 0 because observing $y$ does not change the agent's behaviour or expected return. Shapley values applied to the value function is not capable of capturing this relationship.

\begin{figure}[t]
    \centering
    \includegraphics[width=2.3in]{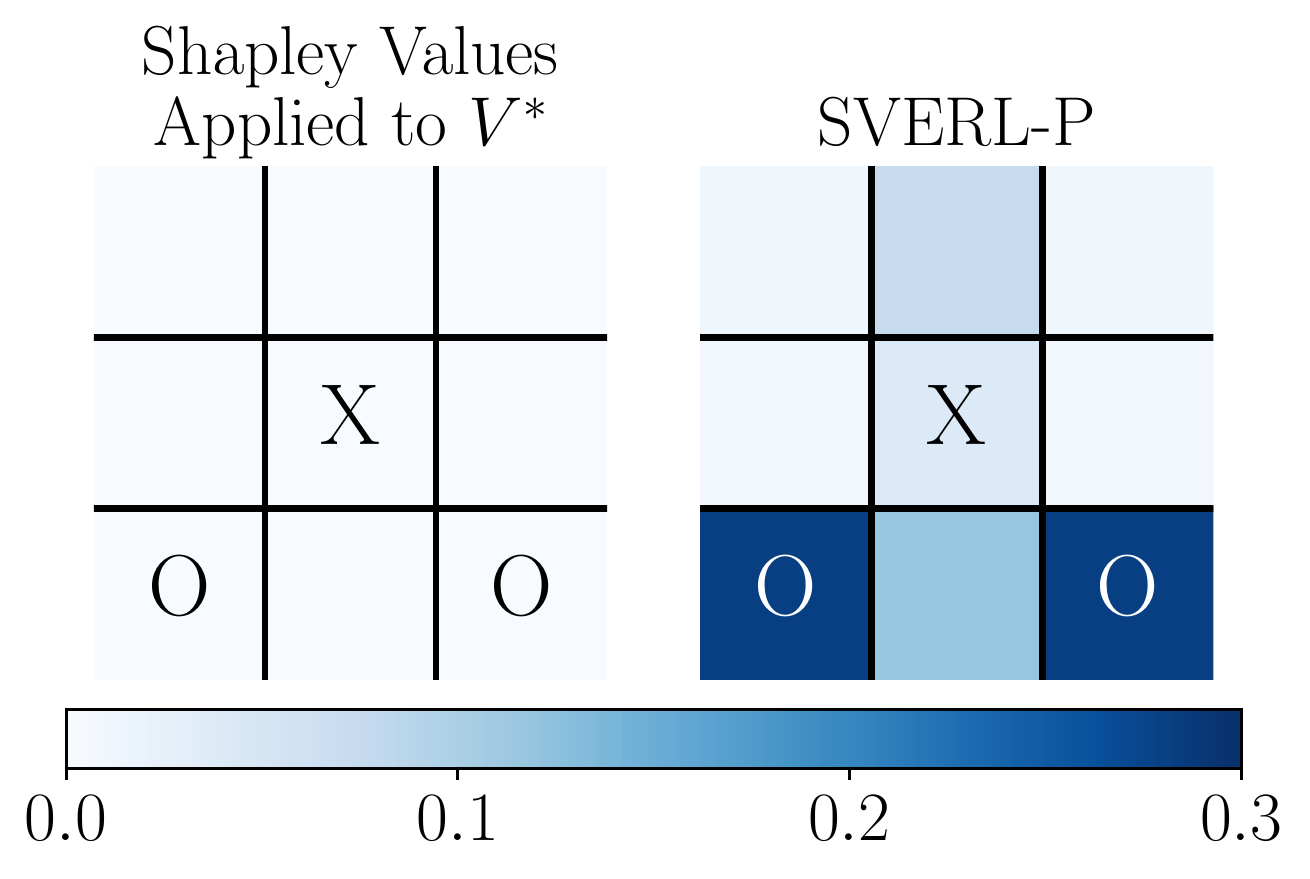}
    \caption{On the left: Shapley values applied to a state-value function in a Tic-Tac-Toe state. On the right: SVERL-P, presented in \cref{sec:SVERL-P}, for the same state. Shapley values are represented using a color scale projected onto each cell. There are 9 state features, corresponding to each position on the board, with possible values \texttt{X}, \texttt{O} or \texttt{empty}. The agent plays as \texttt{X} against opponent \texttt{O}.}
    \label{fig:xando_comparison}
\end{figure}

In Tic-Tac-Toe, there are 9 features, corresponding to each board position, with possible values \texttt{X}, \texttt{O} or \texttt{empty}. Consider an agent (\texttt{X}) that uses $V^*$ to play optimally against an opponent (\texttt{O}) that follows a Minimax policy~\citep{Polak1989}. The reward is $-1$ for losing and 0 for drawing, the only possible outcomes when playing against Minimax. In the state shown in \cref{fig:xando_comparison}, the two squares marked by the opponent inform the agent that it needs to make a blocking move. Intuitively, we would expect the corresponding two state features to impact the performance of the agent. However, the feature contributions identified by applying Shapley values to $V^*$ are zero for every state feature. The reason is that an optimal agent always draws, hence the optimal value function always predicts a return of zero, independently of which state features are observed. These Shapley values explain the value function as a static predictor. They do not consider that the value function depends on the policy, which would change in the absence of some state features.

\textbf{Shapley values applied to policies.} We follow the theory of on-manifold Shapley values in supervised learning to propose the following characteristic value function for a policy $\tilde{\pi} : \S \rightarrow \A$ that outputs actions:
\begin{equation}\label{eq:policyactionshap}
    v^{\tilde{\pi}} \left(\C\right) \defeq \tilde{\pi}_\C(s) = \sum_{s' \in \S}{p^{\tilde{\pi}}(s'|s_\C)\tilde{\pi}(s')},
\end{equation}
and the following characteristic value function for a policy $\pi : \S \cross \A \rightarrow [0, 1]$ that outputs action probabilities:
\begin{equation}\label{eq:policyprobshap}
    v^{\pi}\left(\C\right) \defeq \pi_\C(a | s) = \sum_{s' \in \S}{p^{\pi}(s'|s_\C)\pi(a | s')}.
\end{equation}
\cref{eq:policyactionshap,eq:policyprobshap} account for feature correlations by using the conditional limiting state occupancy distribution $p^{\pi}(s'|s_\C)$, as in \cref{eq:valueshap,eq:qvalueshap}. We note that \cref{eq:policyactionshap} is not valid in discrete action spaces because it is not meaningful to sum discrete actions.

The characteristic value functions in \cref{eq:policyactionshap,eq:policyprobshap} produce Shapley values that show the contribution of state features, respectively, to the action selected by an agent and to the probability of selecting action $a$. Both values provide information on the contributions of state features to the decision made by the agent. This insight is valuable but we argue that there is more to be understood and explained about the decision. Specifically, these Shapley values reveal no insight into the importance of state features for an agent's performance.

As an illustrative example, imagine an agent planning the shortest route through a city. The agent arrives at a junction where turning left and turning right both result in an optimal route. Assume that the agent's policy is to turn left if it observes a road sign (a state feature), and to turn right otherwise. Shapley values applied to the agent's policy would assign a large contribution to the road sign, which is justified and improves our understanding of the agent's behaviour. The sign was indeed instrumental in the agent's decision to turn left. However, one would be incorrect to then conclude that the sign is important for the agent to perform well. On the contrary, because turning left and turning right are both optimal, the sign contributes nothing to the agent's performance. This insight can be gained only by considering the effect of removing state features on the agent's performance. Therefore, we make a distinction between explaining why the agent acted in a specific way and explaining how features impact agent performance.

The contributions of state features to the value function or to the policy do not reveal insight into contributions to agent performance. These two approaches consider either the contributions to predicting expected return independent of behaviour or the contributions to behaviour independent of expected return. We have highlighted the limitations of both approaches. Next we propose an approach to explaining reinforcement learning by identifying contributions of state features to agent performance.

\section{Explaining Agent Performance}
\label{sec:SVERL-P}
Here we provide a formulation of Shapley values to explain the performance of a reinforcement learning agent. We present two methods that explain either the local or the global contributions of state features to performance. Each approach reveals unique insight that improves understanding. In both approaches, state features are removed from an agent's observation for certain states, then the performance of the resulting policy is evaluated using expected return. We call this approach SVERL-Performance (SVERL-P).

\textbf{Local explanations.} Local SVERL-P explains the contributions of state features to performance from state $s$ by considering removing state features from an agent's observation of state $s$. For some policy $\pi : \S \cross \A \rightarrow [0, 1]$ to be explained, the local SVERL-P characteristic value function is given by:
\begin{align}
	v^{\text{local}}{\left(\C\right)} \coloneqq ~\mathbb{E}_{\hat{\pi}}{\left[\sum_{t=0}^\infty\gamma^t r_{t+1} | s_0 = s\right]}, \label{eq:obsstateexpretapen} \\\nonumber\\
	\text{where} ~~ \hat{\pi}(a_t | s_t) =
	\begin{cases}
		\pi_\C\left(a_t|s_t\right) & \text{if } s_t = s, \\
		\pi{\left(a_t|s_t\right)} & \text{otherwise.}
	\end{cases}\nonumber
\end{align}

Shapley values resulting from \cref{eq:obsstateexpretapen} show the contribution of each feature to the change in performance when all state features are observed in state $s$, $v^{\text{local}}{\left(\F\right)}$, compared to when no state features are observed in state $s$, $v^{\text{local}}{\left(\emptyset\right)}$.

In most problems, state features are not independent, so we use the theory for on-manifold Shapley values to propose sampling $a$ from the agent's policy given that it observes $s_\C$:
\begin{equation}\label{eq:policyprobSVERL-P}
    \pi_\C(a | s) = \mathbb{E}_{p^\pi(s'|s_\C)}\left[\pi(a | s')\right],
\end{equation}
where we suggest the conditional data distribution in \cref{eq:onmanifoldexpectation} becomes the conditional limiting state occupancy distribution $p^\pi(s'|s_\C)$.

The resulting explanations are specific to the policy used, which can be any possible policy, including a suboptimal policy. One can interpret $\pi_\C$ as the policy that best tries to match the behaviour of the original policy $\pi$ given that features are missing. Policy $\pi_\C$ will not usually be able to perfectly mimic the behaviour of policy $\pi$. It is exactly this difference in behaviour that causes the change in performance.

\textbf{Global explanations.} Local SVERL-P considers the contributions of state features to performance from a single state. In addition to such local contributions, one may wish to understand the contributions of state features to performance globally. For example, in autonomous driving, a user may wish to understand which parts of an autonomous vehicle's observations are most important for driving performance, to focus resources on improving those parts of the road system. Some state features might contribute substantially to performance in certain states, such as breaking when observing a human or pulling over when an ambulance approaches, while road markings may be globally important by contributing to agent performance in many states.

To quantify the global impact of state features on agent performance, we consider the effect of removing state features from every state in an environment. The corresponding (global) SVERL-P characteristic value function is as follows:
\begin{equation}\label{eq:obsallexpretapen}
	v^{\text{global}}{\left(\C\right)} \coloneqq ~\mathbb{E}_{\pi_\C}{\left[\sum_{t=0}^\infty\gamma^t r_{t+1} | s_0 = s\right]}.
\end{equation}
\cref{eq:obsallexpretapen} produces Shapley values that show the contribution of state features to performance in state $s$ and all future states that follow. These Shapley values are still conditioned on state and therefore not a truly global explanation method. To produce a fully global explanation, one can marginalise over the state space using the limiting state occupancy distribution, producing global SVERL-P:
\begin{equation}\label{eq:glob}
    \Phi_i(v^{\text{global}}) = \mathbb{E}_{p^\pi(s)}{\left[\phi_i\left(v^{\text{global}}, s\right)\right]}.
\end{equation}
\cref{eq:glob} gives the contribution of a state feature to the performance of the agent in its environment. An alternative is to marginalise over the initial state distribution $p_0$, which would place undue attention on the initial states and is therefore less useful in infinite-horizon problems.

\section{Experiments}
\label{sec:exp}
We present experimental results in a variety of domains. We contrast SVERL-P with applying Shapley values to policies and to value functions, demonstrating the limitations of the latter approaches. All Shapley values are calculated exactly, as described in \cref{app:svexact}. The domains are fully described in \cref{app:domains}. 

\textbf{Gridworld-B.} We first consider Gridworld-B, shown in \cref{fig:gwb}. Imagine an agent acting optimally: choosing East (E) in state 1 and North (N) in every other state.

Consider local explanations for specific states. Whatever the state, if neither $x$ nor $y$ is known, the agent cannot know the optimal action with certainty but it knows that the optimal action is either N or E and that N is more likely than E. 

In states 3 and 4, either the $x$ or the $y$ feature is sufficient for the agent to take the optimal action N; in other words, $x$ and $y$ features make an equal contribution to agent  performance. Furthermore, this contribution is rather small because, if neither feature is known, N is still the likely optimal action.  

In state 1, the $x$ feature is sufficient for the agent to take the optimal action E (because an optimal agent is never in state 5). The $y$ feature also improves the agent's performance, but by a smaller amount, because it increases the probability of the agent selecting the optimal action E. In sum, the $x$ and $y$ features contribute positively to agent performance, with the $x$ feature contributing more. 

In state 2, the $x$ feature is sufficient for the agent to take the optimal action N while the $y$ feature actually decreases the probability of selecting the optimal action N (it increases the probability of selecting the suboptimal action E). The $x$ feature therefore makes a positive contribution to agent performance while the $y$ feature makes a negative contribution.

\begin{figure}[t]
    \centering
    \includegraphics[width=3.2in]{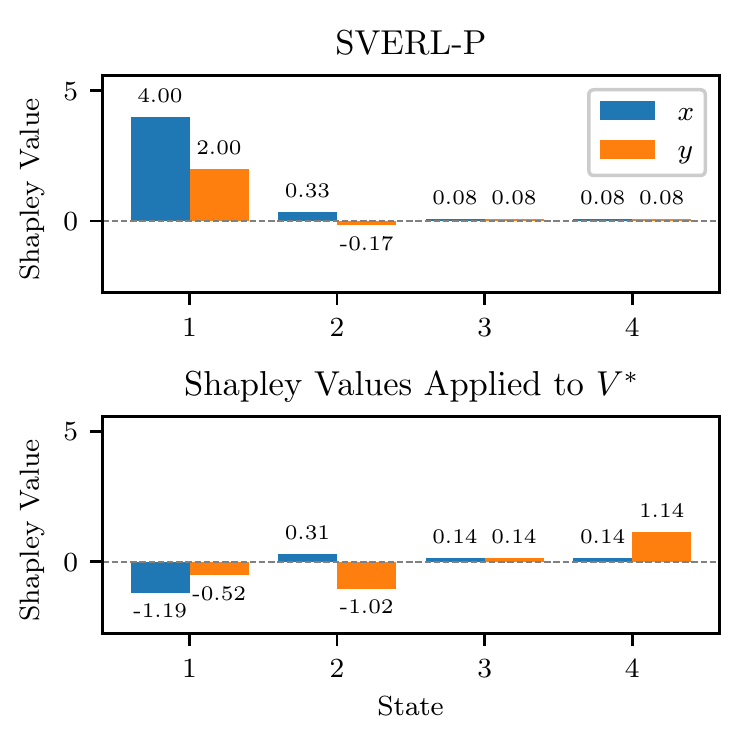}
    \vskip -0.15in
    \caption{Shapley values of $x$ and $y$ state features in Gridworld-B. Top panel: SVERL-P. Bottom panel: Shapley values applied to a value function.}
    \label{fig:gwb_SVERL-P_local}
\end{figure}

Local SVERL-P contributions are shown in the top panel in \cref{fig:gwb_SVERL-P_local}. SVERL-P values align with our intuitive analysis of the domain.  As expected, in states 3 and 4, both $x$ and $y$ contribute a small, equal amount to agent performance. Also as expected, in state 1, $x$ contributes more to performance than $y$. And we can now quantify the difference precisely: $x$ contributes exactly twice as much as $y$. In state 2, Shapley values once again mirror our expectations, with $x$ contributing positively to performance and $y$ contributing negatively\textemdash a reminder that a little bit of knowledge can be a dangerous thing.

Next, consider the global contribution of the two features. Based on the discussion above, the $x$ feature positively contributes larger amounts, more often, to the agent's performance than the $y$ feature. Therefore, we expect the global contribution of the $x$ feature to be larger than that of the $y$ feature. These expectations align with global SVERL-P contributions: $1.43$ for $x$ and $0.64$ for $y$.

SVERL-P has correctly and precisely expressed the local and global contribution of the features $x$ and $y$ to performance. It has done so in more detail and precision than our intuitive expectations, demonstrating the value of SVERL-P even in such a simple domain.

\begin{figure}[t]
\begin{center}
\centerline{\includegraphics[width=\columnwidth]{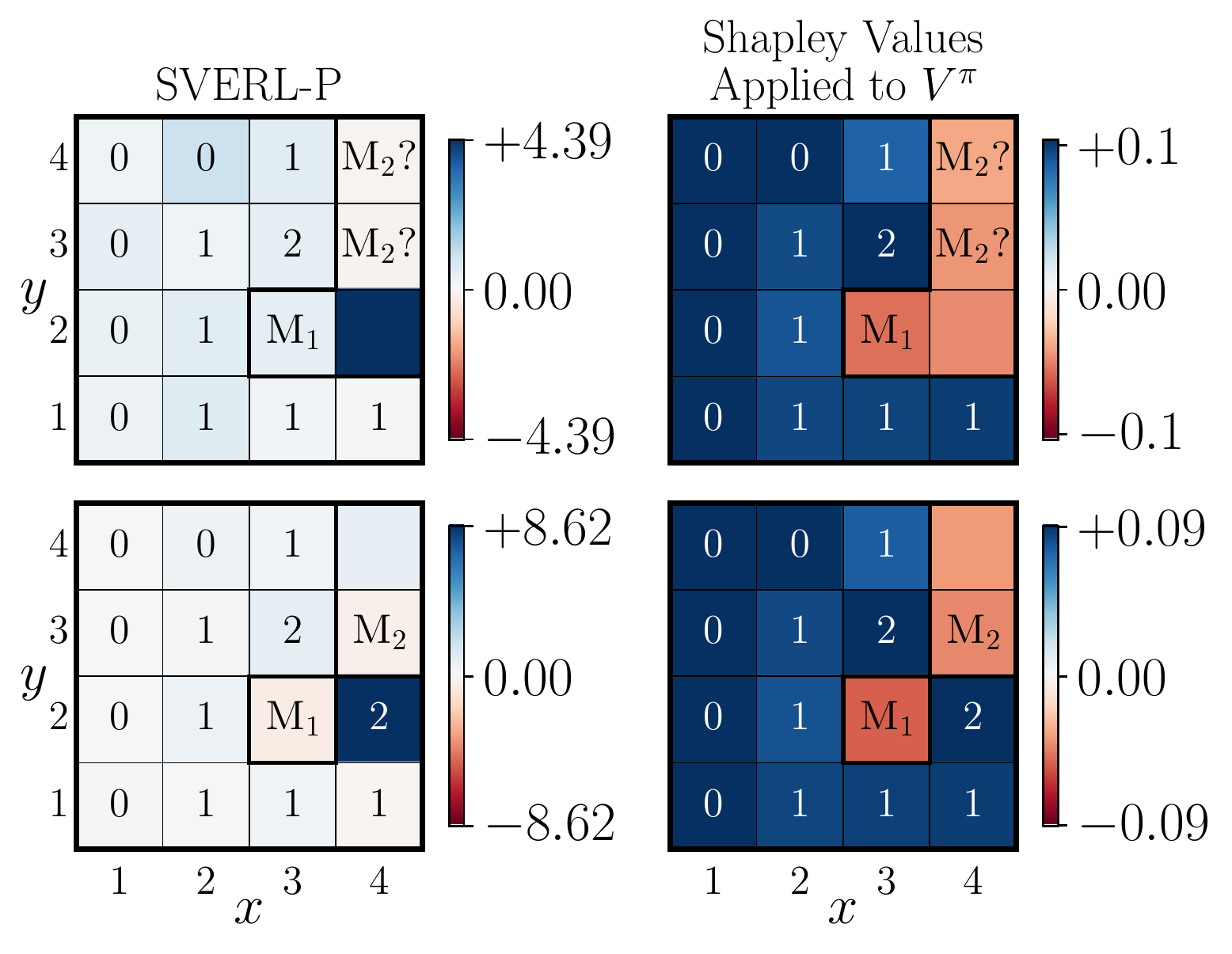}}
\caption{Shapley value contributions for two successive states (top to bottom) of Minesweeper, represented as the color of each cell. On the left: SVERL-P. On the right: Shapley values applied to a value function. The domain contains two mines, hidden from the agent. In the top state, the state features reveal the exact location of one mine and two potential locations for the second mine, marked for reference as ``\textsc{M}$_1$" and ``\textsc{M}$_2$?" respectively. The exact location of the second mine is then revealed in the second state, marked as ``\textsc{M}$_2$" for reference.}
\label{fig:svrl_mine2}
\end{center}
\end{figure}

\textbf{Minesweeper.} This is a relatively large domain, with approximately 175,000 states, where it can be difficult to identify how individual state features contribute to performance by reasoning alone. By using SVERL-P, we find local explanations of performance that reveal novel insight into the two successive Minesweeper states shown in \cref{fig:svrl_mine2}. 

The features in this domain are the 16 grid squares, with possible values 0, 1, 2, or unopened. \cref{fig:svrl_mine2} shows that one feature in particular ($x=4$, $y=2$) contributes substantially to performance in both states, with all other features contributing relatively little in comparison. On further inspection, we see that the feature $(4,2)$ is the \emph{only} feature that can exactly determine the location of \textsc{M}$_2$. On the other hand, many features reveal the exact location of \textsc{M}$_1$. To act optimally, the agent must determine the exact location of \textsc{M}$_2$ so the feature $(4,2)$ is the most important one for completing the episode successfully.

Notice the negative SVERL-P contributions for the squares with possible mines. These are discussed in detail in \cref{app:minesweeper}.

\begin{figure}[!t]
\begin{center}
\raisebox{-0.65\height}{\includegraphics[width=2.6in]{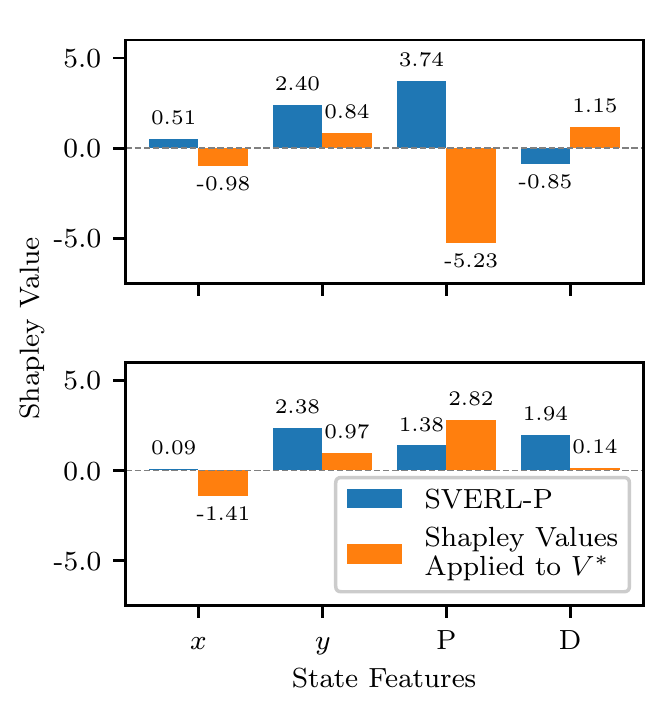}}
\raisebox{-0.65\height}{\includegraphics[width=0.6in]{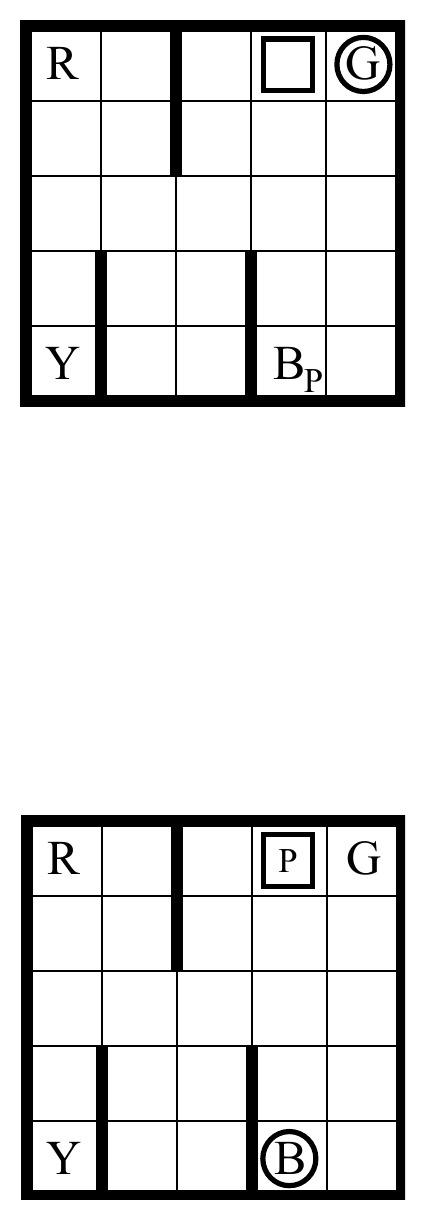}}
\caption{SVERL-P contributions contrasted with Shapley values applied to a value function for two states in the Taxi domain. State features are the $x$ and $y$ coordinates of the taxi, passenger location (P), and destination location (D). The taxi location is marked with a rectangle, the passenger location is marked with a p and the destination location is circled. In the top state, the passenger is at location \texttt{B} and the destination is location \texttt{G}. In the bottom state, the passenger is in the taxi and the destination is location B.}
\label{fig:svrl_taxi}
\end{center}
\vskip -0.2in
\end{figure}

\textbf{Taxi.} In the taxi domain~\citep{Dietterich1998}, the agent picks up a passenger and drops them off at their destination. Rewards are $-1$ for all actions, an additional $+20$ for dropping a passenger at the correct destination, and an additional  $-10$ for attempting to pick up or drop off the passenger at an inappropriate location. We examine the two states shown in \cref{fig:svrl_taxi}. 

In the state shown on the top panel, to successfully complete the episode, the agent must first pick up the passenger. Knowledge of the passenger location is therefore vital and we expect this feature to contribute a large amount to performance. This is captured by SVERL-P, as shown in \cref{fig:svrl_taxi}. Conversely, until the passenger has been collected, we do not expect the destination location to contribute positively to performance. Surprisingly, SVERL-P shows that observing the destination location actually reduces the agent performance. Upon closer review, we see that, in this state, observing the destination location without the passenger location increases the probability of navigating towards the destination, which is a suboptimal action.

SVERL-P also shows that the $x$ feature has a relatively low contribution to performance compared to the $y$ feature. Consider an agent that observes $x=4$ but cannot observe its $y$ coordinate. There are five possibilities for the value of $y$. One of them, $y=1$, would result in executing the pick-up action. Not being able to observe $y$ increases the probability of choosing this action and earning a large negative reward, reducing the agent's expected return. By observing $y$ along with $x$, the agent eliminates the possibility of inappropriate execution of the pick-up action, leading to a large marginal contribution to performance by $y$. Inappropriately executing the pick-up or drop-off action is highly detrimental to performance. Features that decrease this probability are the largest positive contributors to performance.

In the state shown on the lower panel in \cref{fig:svrl_taxi}, the passenger is in the taxi, to be dropped off at location \texttt{B}. The optimal policy navigates to the drop-off location with the passenger in taxi. Intuitively, both the passenger and the destination location are important, as shown by the SVERL-P contributions. The $x$ and $y$ state feature contributions are similar to those in the state discussed previously, for similar reasons\textemdash observing $x$ often increases the probability of inappropriately executing the drop-off action, whereas observing $y$ decreases it.

\textbf{SVERL-P compared with Shapley values applied to value functions.} The domains Gridworld-A and Tic-Tac-Toe were used in \cref{sec:svinrl} to demonstrate that applying Shapley values to an agent's value function does not explain agent performance. In contrast, local SVERL-P contributions in these domains, shown in \cref{fig:value_comparison,fig:xando_comparison}, match our intuitive understanding of the contribution of state features to performance.

\begin{figure}[tb]
\begin{center}
\centerline{\includegraphics[width=2in]{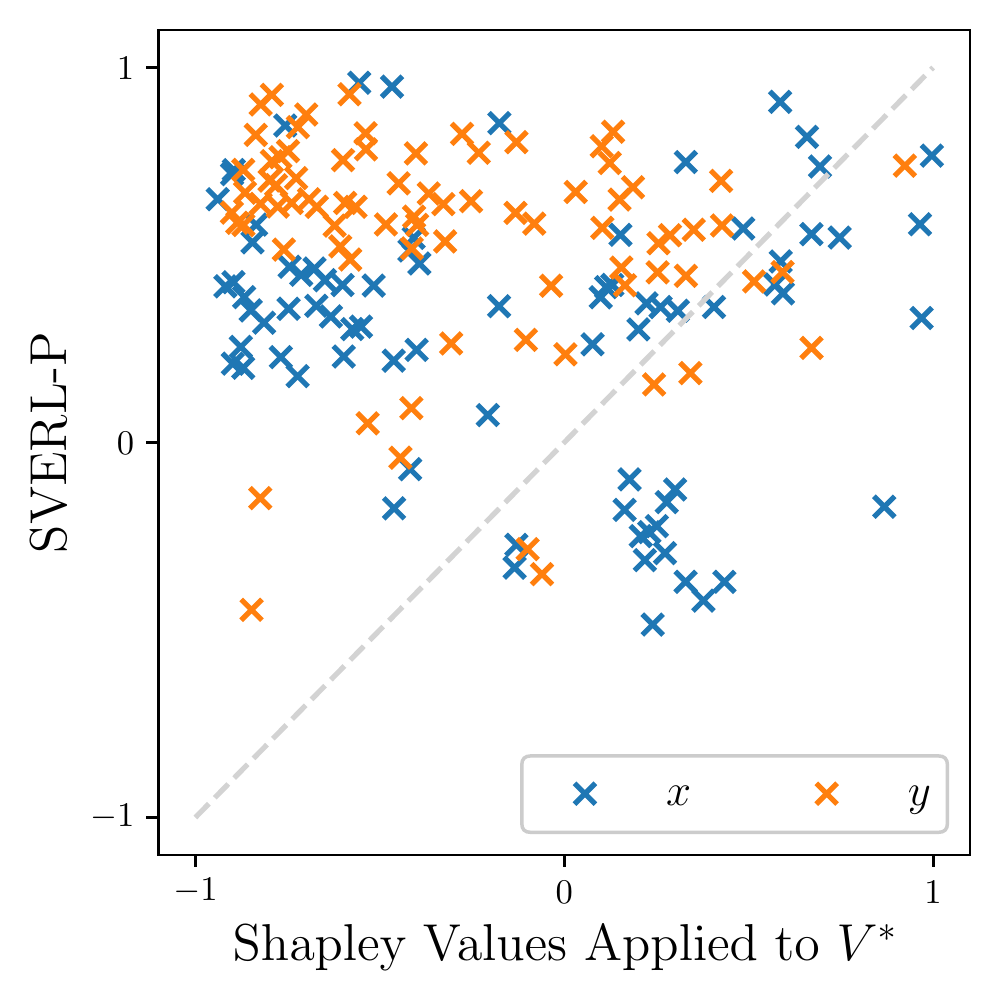}}
\caption{SVERL-P for every state of Gridworld-D compared to Shapley values applied to a value function. Shapley values were normalised to fall between $-1$ and $1$. Each blue cross denotes the $x$ feature for a particular state and each orange cross the $y$ feature.}
\label{fig:SVERL-P_empirical}
\end{center}
\vskip -0.2in
\end{figure}

As a result of purposely choosing simple, illustrative examples, the examples in these two domains used either a constant policy or a constant value function. MDPs with these particular properties are uncommon. Our arguments, however, are valid for any MDP. As an example, \cref{fig:gwb_SVERL-P_local,fig:svrl_mine2,fig:svrl_taxi} show that, in all domains tested, SVERL-P gives different results than applying Shapley values to the value function. They include domains with varying policies and value functions. In \cref{fig:SVERL-P_empirical}, we compare SVERL-P and Shapley values applied to $V^*$ in every state of a randomly-constructed gridworld with 80 states (Gridworld-D). The results show a persistent difference between these two approaches.

\textbf{SVERL-P compared with Shapley values applied to \mbox{policies}.} In \cref{sec:svinrl}, we introduced Shapley values applied to an agent's policy. We argued that they provided insight which improved understanding of a decision but that further insight could be drawn by also considering the effect of state features on performance. We now illustrate our viewpoint by comparing local SVERL-P to Shapley values applied to a policy. 

Consider Gridworld-C, shown in \cref{fig:gwc}. In this domain, if no state feature is known, the agent cannot know the optimal action with certainty but it knows that (1) it is either North, East or West, and (2) North is more likely than East or West. In states 2 and 5, neither observing $x$ nor observing $y$ reveals the optimal action. We have no natural intuition on the importance of state features and must rely on Shapley values.

\begin{figure}[t]
    \centering
    \includegraphics[width=3.2in]{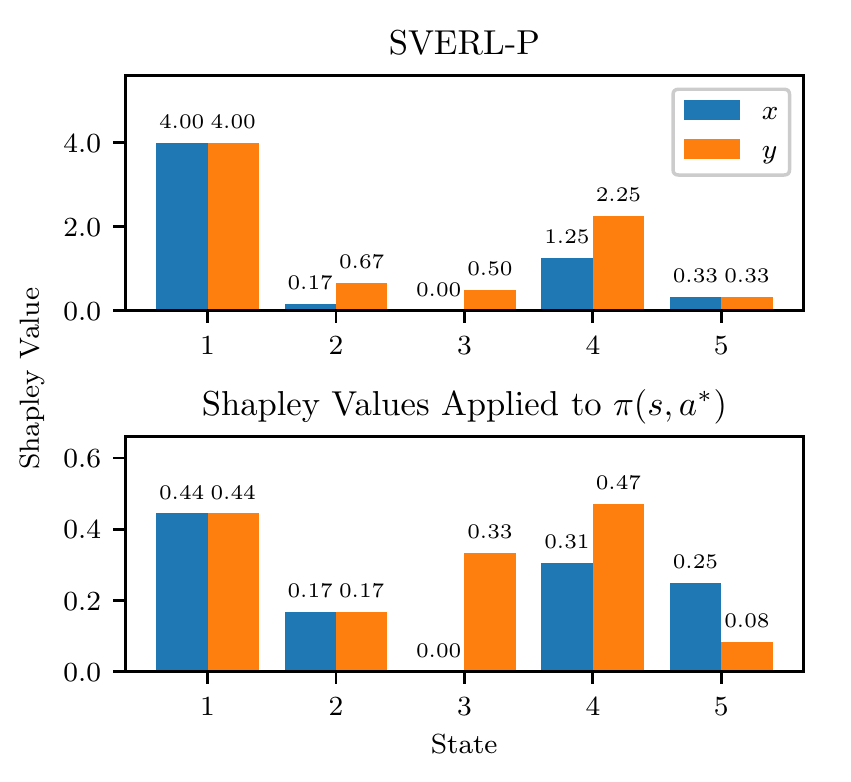}
    \caption{SVERL-P compared to Shapley values applied to a policy in Gridworld-C (\cref{fig:gwc}). The plots show the Shapley values of the $x$ and $y$ state features for all states. SVERL-P gives the contribution of state features towards performance while Shapley values applied to a policy give the contributions of state features towards the likelihood of selecting the optimal action in each state.}
    \label{fig:policy_comparison}
\end{figure}

Shapley values applied to the optimal policy in every state are shown in \cref{fig:policy_comparison}. For each state, the Shapley values are presented for the optimal action, $a^*$. In state 5, $x$ contributes more than $y$ to the probability of choosing the optimal action (N). One might assume that $x$ is therefore more important than $y$ for an agent to act optimally. However, this would be incorrect. The local SVERL-P contributions, shown in the top panel of \cref{fig:policy_comparison}, reveal that in fact the $x$ and $y$ features contribute equally to performance. The reason for this difference is that, in state 5, $x$ also contributes towards the likelihood of selecting the worst action (E). Similarly, in state 2, Shapley values applied to the policy show that both $x$ and $y$ contribute equally to the probability of selecting the optimal action (N). However, local SVERL-P contributions reveal that $y$ actually contributes more than $x$ to performance. In this state, observing $x$ but not $y$ increases the probability of selecting the worst action (W).

By applying Shapley values to policies without considering the consequence on performance, one would draw incorrect or incomplete conclusions about the importance of state features. By considering the contribution of state features towards performance, SVERL-P provides additional insight into agent behaviour.

\section{Discussion}
\label{discussion}

We presented a theoretical and empirical analysis of using Shapley values for explaining reinforcement learning (SVERL), starting from first principles, and demonstrated the limitations of existing work. We then developed SVERL-P, a method that uses Shapley values to explain agent performance. SVERL-P considers the consequences of removing features by explicitly deriving an agent's policy and quantifying the change in performance. Our results show that SVERL-P produces meaningful explanations in a variety of reinforcement learning problems, matching and supplementing human intuition.

In most real-world applications, it is computationally expensive to calculate the SVERL-P characteristic value functions exactly. So the characteristic value functions, and hence the Shapley values, must be approximated. Here we outline an approximation algorithm for local SVERL-P based on the on-manifold sampling approach from Shapley values in supervised learning, which has been proven to converge to the Shapley value in the limit \citep{Strumbelj2010,Frye2020a}. Analogous to the sample in Equation \ref{eq:offmanifoldapprox}, each sample in the algorithm is a marginal gain:
\begin{align*}
    \mathbb{E}_{\pi_1}{\Bigg[\sum_{t=0}^\infty\gamma^t r_{t+1} | s_0 = s \Bigg]} -{}& \mathbb{E}_{\pi_2}{\Bigg[\sum_{t=0}^\infty\gamma^t r_{t+1} | s_0 = s\Bigg]},\\
	\text{where} ~~~~~ \pi_1(a_t | s_t) ={}& 
	\begin{cases}
		\pi{\left(a_t|s'\right)} {}&\text{if } s_t = s, \\
		\pi{\left(a_t|s_t\right)} {}& \text{otherwise,}
	\end{cases}\\
	\text{and} ~~~~~ \pi_2(a_t | s_t) =&
	\begin{cases}
		\pi{\left(a_t|s''\right)} {}& \text{if } s_t = s, \\
		\pi{\left(a_t|s_t\right)} {}& \text{otherwise}.
	\end{cases}
\end{align*}
A new $s'$ is sampled from $p^{\pi}(\cdot|s_{\C\cup\{i\}})$, and a new $s''$ is sampled from $p^{\pi}(\cdot|s_\C)$, whenever $s_t=s$. Each coalition $\C\subseteq \F\setminus \left\{i\right\}$ is sampled proportional to the multinomial term in the Shapley value calculation. The expected returns can be evaluated using a standard reinforcement learning method, such as Monte Carlo rollouts. This sampling method requires the learning of state occupancy distributions $p^{\pi}(\cdot|s_\C)$ for all $\C \subseteq \F$, which is not trivial. We suggest taking inspiration from one of the on-manifold sampling methods proposed by \citet{Frye2020a}. Importantly, it is likely that these distributions do not need to be learnt exactly because optimal policies usually visit only a small subset of states in large domains.

SVERL is a direct application of Shapley values using specific characteristic value functions suitable for reinforcement learning. All the theoretical guarantees of Shapley values apply to SVERL. Similarly, any advancements in applying Shapley values to supervised learning will apply directly to SVERL. For example, SVERL might be difficult to interpret in domains with thousands of features, such as robotics or vision. However, a method such as \emph{groupShapley} \citep{Jullum2021}, which finds the contribution of groups of features and was developed for supervised learning, could be applied to SVERL, offering computational advantages and simplifying interpretation.

As with any feature-based explanation method, there is further work, often psychological and sociological, to derive useful explanations which improve a user's understanding. It is naturally human to interpret Shapley values subjectively, often developing beliefs and understanding that extend beyond the quantitative information that they provide. These interpretations will likely become more challenging and subjective as the number of features increases. When one proceeds to develop this extended understanding, before acting on it, they must first evaluate whether it is well founded. For example, SVERL-P values allow us to say ``this feature contributed $x$ amount to an agent's performance". One can hypothesise on why that feature contributed $x$ but such hypotheses must be tested. These tests depend on the task, explanation and hypothesis. We suggest that future research focuses on (1) the presentation, interpretation and explanatory use of feature attribution techniques such as Shapley values, and (2) methods for evaluating the conclusions drawn from such interpretations. We provide an example in \cref{app:minesweeper}.

\section*{Acknowledgements}

This work was supported by the UKRI Centre for Doctoral Training in Accountable, Responsible and Transparent AI (ART-AI) [EP/S023437/1], the EPSRC Centre for Doctoral Training in Digital Entertainment (CDE) [EP/L016540/1] and the University of Bath. This research made use of Hex, the GPU Cloud in the Department of Computer Science at the University of Bath. We thank our reviewers for a constructive process and the members of the Bath Reinforcement Learning Laboratory for their feedback. We thank Scarllette Ellis for her Minesweeper implementation.

\bibliography{shapley_rl}
\bibliographystyle{icml2023}

\newpage
\appendix
\onecolumn

\section{Domains}
\label{app:domains}

\textbf{Gridworld-A}, shown in \cref{fig:gwa}, is a deterministic gridworld. The MDP state represents the grid square occupied by the agent and is described by two features, $(x, y)$, the $x$ and $y$ coordinates of the agent on the grid. There are six states, $\S = \{(1, 1), (1, 2), (1, 3), (2, 1), (2, 2), (2, 3)\}$, two of which are goal states, $G = \{(1, 3), (2, 3)\}$. The initial state is sampled randomly from the southernmost squares, $\{(1, 1), (2, 1)\}$. The actions are North, East, South, and West. Reward is $-1$ for every action taken and an additional $+10$ for transitioning into a goal state, producing a shortest path problem. Actions that attempt to transition an agent out of the grid do not change the state. \textbf{Gridworld-B}, shown in \cref{fig:gwb}, and \textbf{Gridworld-C}, shown in \cref{fig:gwc}, are identical to Gridworld-A in all aspects other than the grid layout and the identity of the goal states.

\textbf{Gridworld-D} is a deterministic $10 \times 10$ gridworld, containing 20 grid positions that are impassable blocks, selected uniformly randomly from among all grid positions. There is a single goal state, selected randomly, and fixed across episodes. The initial state is selected randomly from among grid squares that are not impassable blocks or the goal. The domain is identical to Gridworld-A in all other aspects.

\textbf{Tic-Tac-Toe} is a classic game played on a $3 \cross 3$ grid, where two players take turns to place noughts (\texttt{O}) and crosses (\texttt{X}). When a player places three noughts or three crosses such that a straight line can be drawn through them, the game ends with a win for the corresponding player. If the grid is full with no winner, the game is a draw. The state has nine features, with each feature representing a specific grid position, taking on values \texttt{X}, \texttt{O}, or \texttt{empty}. The agent plays as \texttt{X} and the opponent as \texttt{O}. The players have equal probability of playing first. The opponent's policy is the Minimax algorithm~\citep{Polak1989}. Optimal play against this opponent ends in a draw.

\textbf{Taxi} is a classic reinforcement learning domain by \citet{Dietterich1998}. We used the implementation by OpenAI Gym~\citep{Brockman2016}. The domain has a grid with four locations, marked {R}(ed), {G}(reen), {B}(lue) and {Y}(ellow). There are four state features: $x \in \{1, 2, 3, 4, 5\}$, $y\in \{1, 2, 3, 4, 5\}$, passenger-location~$\in \{\textrm{R}, \textrm{G}, \textrm{B}, \textrm{Y}, \textrm{in-taxi}\}$, and destination~$\in \{\textrm{R}, \textrm{G}, \textrm{B}, \textrm{Y}\}$. State features $x$ and $y$ represent the taxi's location. Initial taxi location and destination are selected uniformly randomly. For an episode to terminate successfully, the taxi must navigate to the passenger location, pick-up the passenger, navigate to the destination, and drop-off the passenger. At the beginning of an episode, the passenger location is randomly selected among R, G, B, and Y. Once the passenger has been collected, passenger location becomes in-taxi. The actions are north, south, east, west, pick-up, and drop-off. Pick-up action successfully picks up the passenger only when the taxi and the passenger is at the same grid location. Similarly, drop-off action successfully drops off the passenger when the passenger is in the taxi and the taxi is at the destination. The reward is $-1$ for each action, an additional $+20$ for delivering the passenger at the destination, and $-10$ for unsuccessful execution of the pickup or the drop-off action.

\textbf{Minesweeper} is an implementation of the classic game on a $4 \cross 4$ grid. Each episode resets a grid that contains two hidden mines, each placed randomly. The state has 16 features, with each feature respresenting a specific grid square, taking on values 0, 1, 2 or unopened. Initially, all grid squares are unopened. At each decision stage, the agent selects an unopened square to reveal what is underneath. If it happens to be a number, that number represents the total number of mines in the (up to eight) squares directly surrounding the newly opened square. If the number is zero, all surrounding grid squares are recursively revealed to reveal an area of zeros bordered by strictly positive numbers. The game ends when the agent opens a square with a mine or all squares that do not contain a mine are opened. There is only one reward signal: $-20$ whenever the agent reveals a mine. Therefore the highest return possible is $0$. There is no incentive for the agent to complete a game in minimal time. 

\section{Extended Analysis in Minesweeper}\label{app:minesweeper}

In the minesweeper example of \cref{fig:svrl_mine2}, SVERL-P contributions are negative for two unopened squares (\textsc{M}$_1$ and \textsc{M}$_2$) in the second state. The implication is that observing either state feature makes a negative contribution to the expected return. We hypothesise that, by becoming observable, these features increase the probability that the agent clicks on the corresponding squares. Such an action would reveal the underlying mine and terminate the game with a large negative reward. 

\begin{figure}[!t]
    \centering
    \includegraphics[width=4.5in]{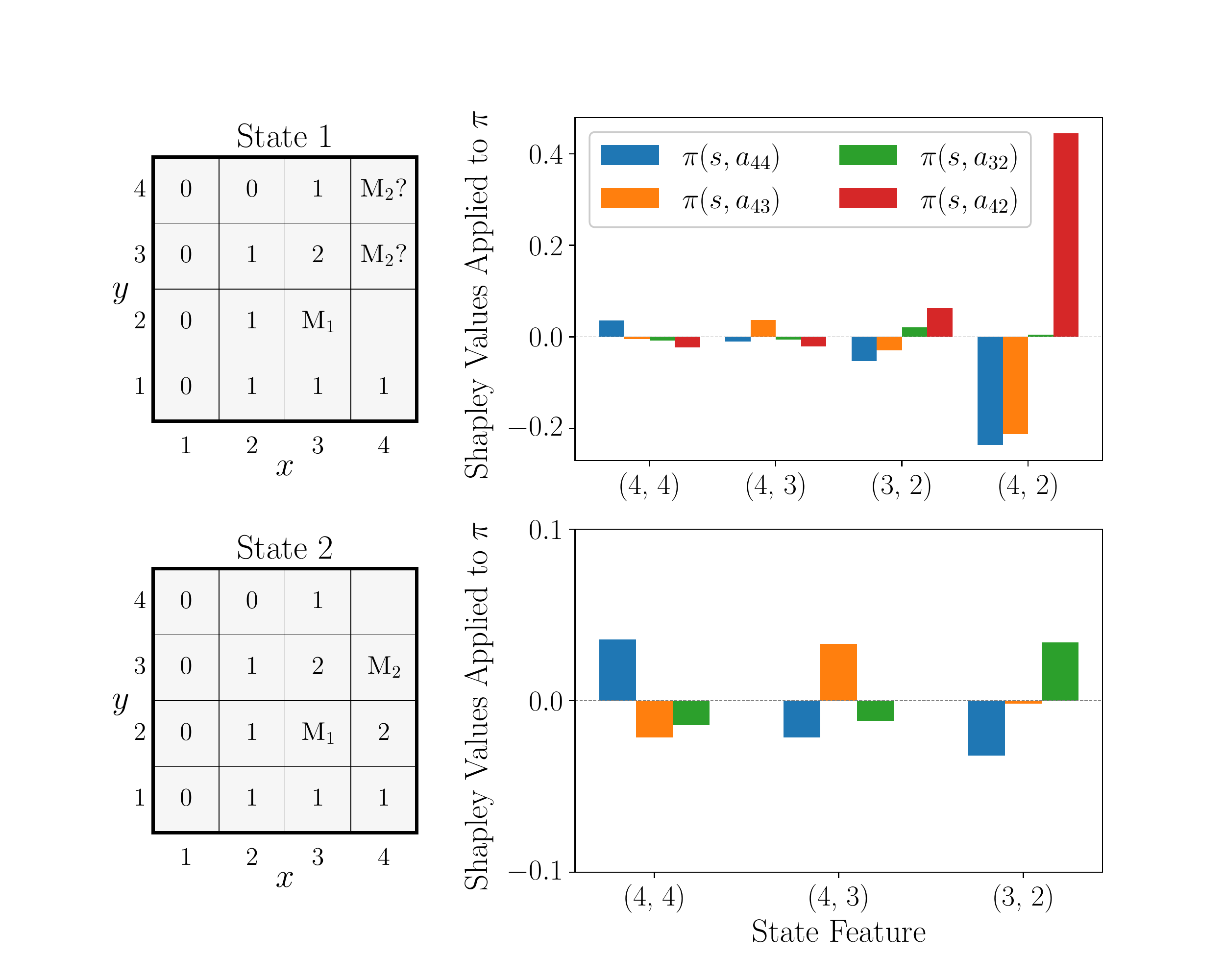}
    \vskip -0.2in
    \caption{Shapley values applied to a policy in two states of Minesweeper. Action $a_{xy}$ denotes the action that opens grid square ($x$, $y$). The plots show, for each available action, the Shapley values of the state features that correspond to unopened squares.}
\label{fig:svrlb_mine}
\end{figure}

In Section~\ref{discussion}, we suggested that humans are likely to naturally over-interpret SVERL-P contributions, developing hypotheses that must be tested. This is one such example. The validity of our hypothesis can be tested by examining Shapley values applied to a policy that outputs action probabilities, introduced in \cref{sec:svinrl}. \cref{fig:svrlb_mine} shows that the Shapley values for the probability of selecting each unopened feature are positive, showing that, on average, observing that a square is unopened positively contributes towards the probability of selecting it.

Note the non-negative SVERL-P contribution of \textsc{M}$_1$ in state 1 even though observing that a square is unopened increases the probability of opening it. On closer inspection, \cref{fig:svrlb_mine} reveals that observing that square $(3, 2)$ is unopened increases the probability of opening square $(4, 2)$ (the optimal action) much more than it increases the probability of opening $(3, 2)$.

SVERL-P contributions revealed insight into \emph{how} features contributed to performance but further analysis was required to investigate \emph{why} features contributed to performance.

\section{Computing Shapley Values}\label{app:svexact}

This work presented four applications of Shapley values in reinforcement learning, under the SVERL framework: Shapley values applied to value functions, Shapley values applied to policies, local SVERL-P and global SVERL-P. Each of the different Shapley values are computed using \cref{eq:sv}, with their respective characteristic value functions computed using \cref{eq:valueshap,eq:qvalueshap,eq:policyprobshap,eq:policyactionshap,eq:obsstateexpretapen,eq:obsallexpretapen}. All of these characteristic value functions require the conditional limiting state occupancy distributions, $p^\pi(s'|s_\C)$, for every $\C \subset \F$. We calculate each $p^\pi(s'|s_\C)$ using Bayes's rule:
\begin{equation} \label{eq:pomdpheuristicsimple2}
	p^{\pi}(s'|s_\C) = \frac{p(s_\C|s')p^{\pi}(s')}{p^{\pi}(s_\C)} 
	= \frac{p(s_\C|s')p^{\pi}(s')}{\sum_{s' \in \S}{p(s_\C|s')p^\pi(s')}}, 
\end{equation}
where the limiting state occupancy distribution $p^\pi(s')$ is approximated through interaction with the environment. Additionally, if $s_\C$ is a possible observation of $s'$, then $p(s_\C|s') = 1$, else $p(s_\C|s') = 0$. For example, in Gridworld-B, $s_\C=\{x=1\}$ is a possible observation of $s'=\{x=1, y=3\}$, whereas $s_\C=\{x=2\}$ is not.

After computing the conditional limiting state occupancy distributions using \cref{eq:pomdpheuristicsimple2}, the characteristic value functions for Shapley values applied to policies and Shapley values applied to value functions can be calculated directly using  \cref{eq:valueshap,eq:qvalueshap,eq:policyprobshap,eq:policyactionshap}. For the local and global SVERL-P characteristic values in \cref{eq:obsstateexpretapen,eq:obsallexpretapen}, first $\pi_\C(a|s)$ must be computed using \cref{eq:policyprobSVERL-P}. Then the characteristic values, which are expected returns, can be computed using any standard reinforcement learning algorithm. We used Monte Carlo roll outs.

\section{Code}

Code is available at \url{https://github.com/bath-reinforcement-learning-lab/SVERL_icml_2023}.

\end{document}